\documentclass[a4paper, 10pt, conference]{ieeeconf}      
\usepackage{FG2021}

\usepackage{times}  
\usepackage{helvet} 
\usepackage{courier}  
\usepackage[hyphens]{url}  
\usepackage{graphicx} 
\usepackage{multirow}
\usepackage{subcaption}
\usepackage{amssymb}
\usepackage[switch]{lineno}

\FGfinalcopy 

\IEEEoverridecommandlockouts                              
\def\FGPaperID{78} 

\title{\LARGE \bf
Harnessing Geometric Constraints from Emotion Labels to\\Improve 
Face Verification
}

\author{\parbox{16cm}{\centering
    {\large Anand Ramakrishnan, Minh Pham and Jacob Whitehill}\\
    {\normalsize
    Worcester Polytechnic Institute}}
}

\begin{document}

\ifFGfinal
\thispagestyle{empty}
\pagestyle{empty}
\else
\author{Anonymous FG2021 submission\\ Paper ID \FGPaperID \\}
\pagestyle{plain}
\fi
\maketitle

\begin{abstract}
For the task of face verification, we explore the utility of harnessing auxiliary facial emotion labels to impose explicit geometric constraints on the embedding space when training deep embedding models. We introduce several novel loss functions that, in conjunction with a standard Triplet Loss \cite{schroff2015facenet}, or ArcFace loss \cite{deng2019arcface}, provide geometric constraints on the embedding space; the labels for our loss functions can be provided using either manually annotated or automatically detected auxiliary emotion labels. 
Our method is implemented purely in terms of the loss function and does not require any changes to the neural network backbone of the embedding function.
\end{abstract}

\section{INTRODUCTION}
\begin{figure*}[!htb]
        \centering
        \begin{subfigure}[b]{\columnwidth}  
            \centering 
            \includegraphics[trim={.0in 1.in .0in 1.1in},clip,width=\textwidth]{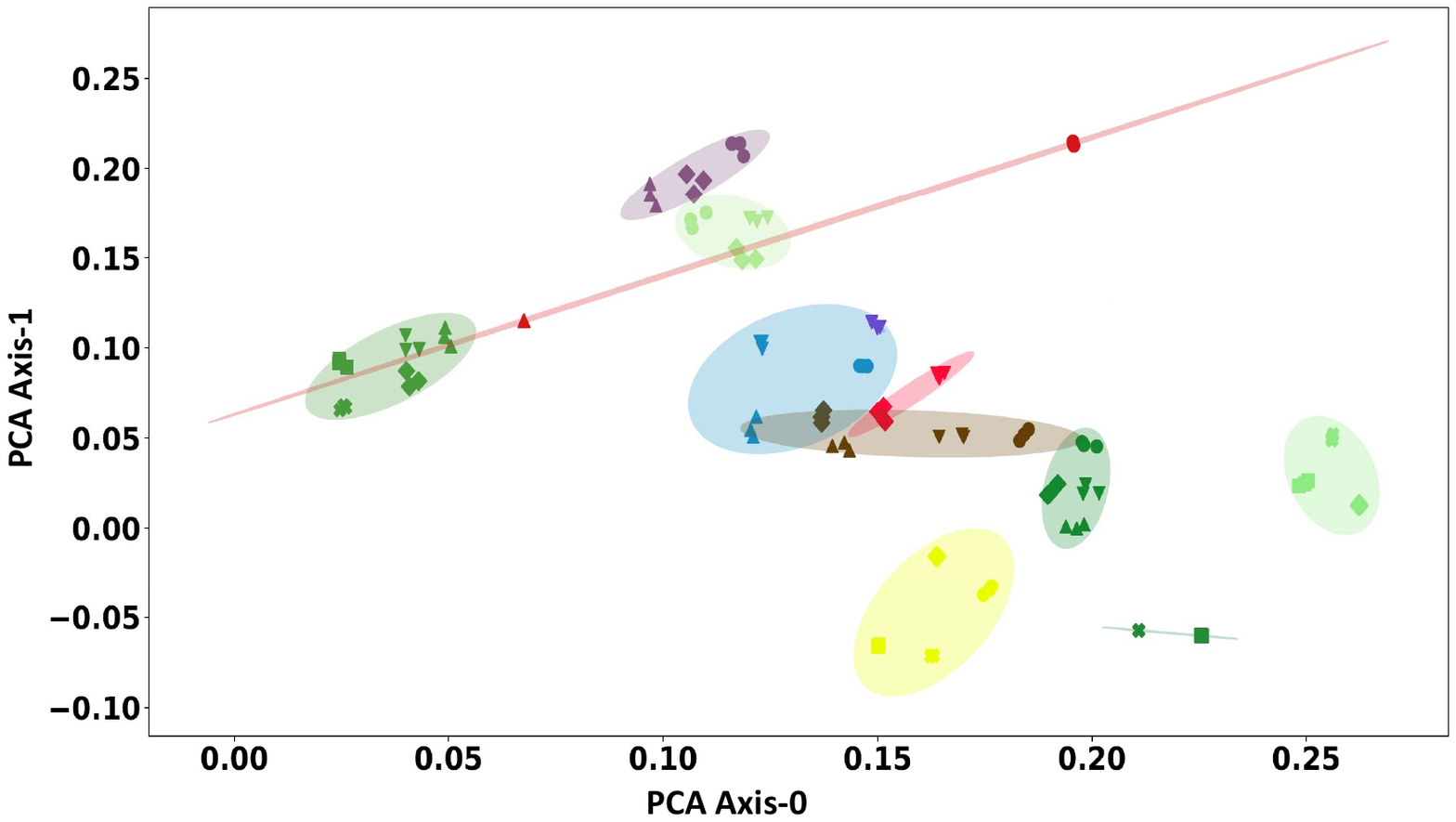}
            \caption[]%
            {{\small Embeddings trained with standard Triplet Loss.}}    
            \label{fig:mean and std of net24}
        \end{subfigure}
        \begin{subfigure}[b]{\columnwidth}
            \centering
            \includegraphics[trim={.0in 1.in .0in 1.1in},clip,width=\textwidth]{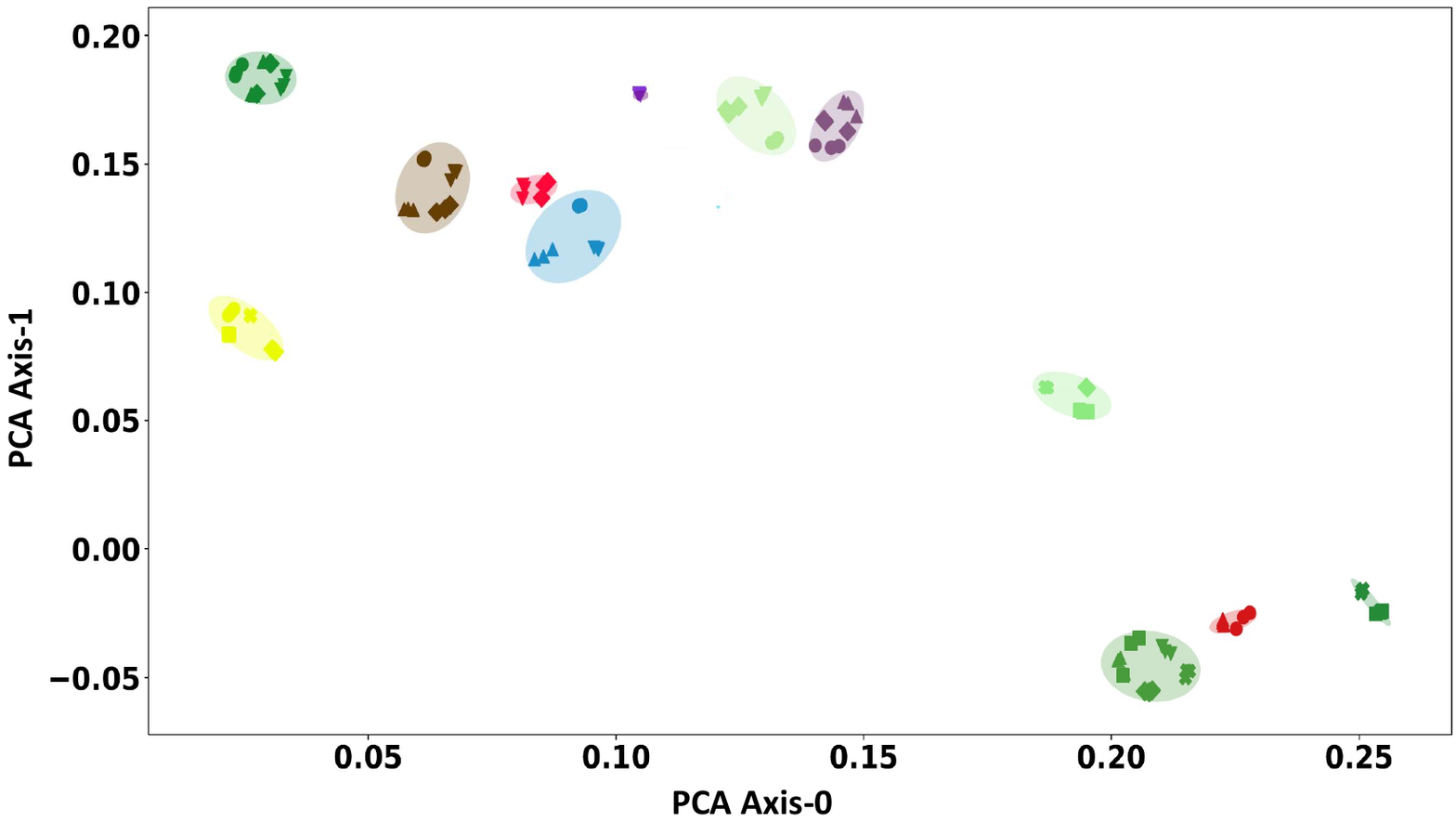}
            \caption[Network2]%
            {{\small Embeddings with geometric constraints  on auxiliary labels.}}    
            \label{fig:mean and std of net14}
        \end{subfigure}
        \caption[ The average and standard deviation of critical parameters ]
        {PCA of the embeddings of the test data after training with (a) Triplet Loss, or (b) a combination of Triplet Loss and two additional novel loss functions ($L_\textrm{PDP}+L_\textrm{FBV}$) that harness the auxiliary emotion labels. Color represents the one-shot class (face ID), and shapes represent the auxiliary information (facial expressions). 
        By training with these extra loss terms, we hope that the clusters corresponding to one-shot classes  become more regular and better separated.} 
        \label{fig:PCA}
    \end{figure*}
    
The goal of \emph{face verification} is to determine automatically whether two face images belong to the same person. The state-of-the-art algorithms for face verification are based on embedding models that use deep convolutional neural networks (e.g., \cite{schroff2015facenet,deng2019arcface,wang2018cosface,liu2017sphereface}) to map face images into an embedding space such that multiple embeddings from the same person are close together and embeddings from different people are far apart. Deep embedding models can be used to measure the similarity between two inputs, even when they come from classes not seen during training. 
Current research in face verification is diverse, including equity in face verification accuracy \cite{nagpal2019deep,wang2020mitigating,ryu2017inclusivefacenet},  multimodal representations \cite{al2018multimodal,al2020multi,shyam2015identifying}, 3-dimensional face verification, partial face verification \cite{lei2016two,elmahmudi2019deep,he2018dynamic,duan2018topology}, loss function design \cite{hadsell2006dimensionality, schroff2015facenet,hermans2017defense,deng2019arcface}, large-scale datasets \cite{cao2018vggface2,guo2016ms,chung2018voxceleb2}, and network architectures \cite{cao2018pose,deng2019retinaface}.

{\bf Contribution}:
Our paper focuses on \emph{harnessing auxiliary emotion label information} to construct an embedding space that can better explain the intra-class variance and thus better separate different classes (i.e., face IDs). Could training with these auxiliary labels result in a more accurate face verification system? To date,  few works have investigated this question.  In particular, we propose several novel geometry-constraining loss functions to encourage structure in the alignment of examples with the same face emotion labels by explicitly using either manually annotated or automatically detected emotion information. 


{\bf Notation}:
Let each example (face) be denoted by $x$. The embedding function $f$ maps $x$ into an embedding vector $y$. The one-shot class label (identity of the person) of $x$ is denoted by $c(x)$, and the auxiliary label (facial expression) of $x$ is denoted by $e(x)$.

\section{Related Work}
\subsubsection{Multi-Task Learning}
One prominent method of using auxiliary labels to improve generalization and obtain better latent representations is multi-task learning (MTL),
an active field of literature for over 20 years \cite{caruana1997multitask}. Learning multiple tasks using a shared representation helps to regularize the model and improve its ability to generalize \cite{ruder2017overview}. MTL has been successfully used in wide array of spaces including natural language processing \cite{collobert2008unified}, object detection \cite{girshick2015fast,ren2015faster} and in drug discovery \cite{ramsundar2015massively}. MTL can be interpreted as providing \emph{implicit} structure on the embedding space to encourage better generalization; in contrast, our methods impose \emph{explicit} structure on where different mini-clusters (corresponding to different auxiliary labels) within each cluster (face ID) should be located.

\subsubsection{Self Supervised Learning}
A promising new approach to learning representations of faces and other images is based on self-supervised learning \cite{wiles2018self, sharma2019self}. 
Self-supervision uses different proxy losses that enable it to learn suitable features for various downstream tasks. This is similar to an MTL setting, except the labels for the proxy losses are obtained through symmetry in the input data (e.g., the embeddings of face images from consecutive frames in a video should be very similar). Like MTL, the constraints imposed by self-supervision on the embedding space are \emph{implicit}. In contrast, the loss functions we impose \emph{explicit} constraints on the embedding space using auxiliary labels from either a human annotator or a pre-trained classifier of the auxiliary attributes.

\subsubsection{Auxiliary labels to improve embedding spaces}
\cite{deng2019retinaface} showed that including a wide variety of prediction tasks such as facial keypoint detection, face detection, etc., improves accuracy for the primary task of face verification. Fusing auxiliary information can help in speaker verification, as seen in \cite{toshniwal2017multitask}. 
\cite{rudolph2017structured} impose hierarchical priors using auxiliary labels to improve exponential-family embeddings and help in the primary task of capturing changes in word usage across different domains. The idea of hierarchical clusterings in the embedding space also inspired the PDM method presented in section \ref{sec:PDM}. 
The work most similar to our own is by \cite{tsai2017improving}, who propose a kernel-based constraint between image representations and auxiliary information. The authors obtain different auxiliary information (word embeddings, human annotations, etc.) and use deep kernel learning to construct an affinity kernel. The authors propose to maximize the relationship between the learned kernel and the corresponding embedding for the data. The PDP loss function proposed in section \ref{sec:PDP} is a looser form of this method.

\subsubsection{Euclidean vs. Spherical Embeddings}
Most prominent deep embedding models \cite{schroff2015facenet,deng2019arcface,yi2014deep,banerjee2005clustering,gopal2014mises,vinyals2016matching} map their inputs into a hypersphere embedded in $n$-dimensional Euclidean space; these have the advantage that computing cosine similarity is trivial and that the distances between vectors are determined just based on their direction, not their lengths. However, embeddings into unconstrained Euclidean space are also possible and have been explore  in the fields of word embeddings \cite{mikolov2013efficient}, image retrieval \cite{oh2016deep,sohn2016improved} and face verification \cite{parkhi2015deep,wen2016discriminative}.
In our work, we explore both spherical and Euclidean embeddings.

\subsubsection{Compositional Embeddings}
Recently there has been an interest in training neural networks to perform set operations (e.g., union) among multiple vectors in an embedding space to reflect higher-order relationships. Much of this work is for word embeddings  \cite{pollack1989implications,nakov2019semeval,lake2018generalization}. However, a few works have also analyzed how embeddings of images with different class labels can be composed for multi-label one-shot learning. Both \cite{li2020compositional} and \cite{alfassy2019laso} present a compositional embedding model that can perform a different set of operations such as ``contains'', ``union'', etc., and achieves higher accuracy in multi-label one-shot learning tasks than traditional embedding methods. In one formulation \cite{li2020compositional}, two embedding functions $f$ and $g$ are trained jointly: $f$ embeds an example (e.g., a face image) into the embedding space, whereas $g$ maps from the embedding space to itself, to preserve certain relationships. While these methods utilize compositional models for multi-label one-shot learning,  we extend this line of work by investigating how training $f$ and $g$ jointly can impose useful geometric constraints on the embedding space and help the model achieve higher accuracy for single-class one-shot learning.

\begin{figure*}
    \centering
    \includegraphics[width=\textwidth]{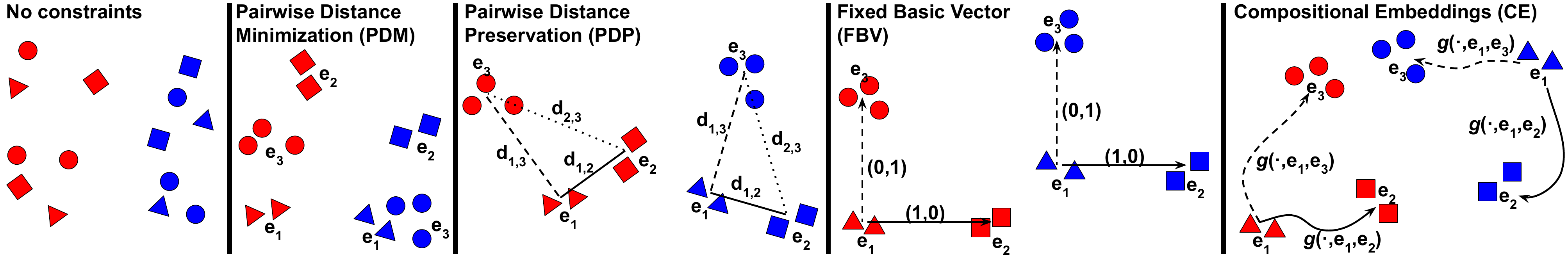}
    \caption{The  geometric constraints on auxiliary labels that we explore  to improve face verification. Colors are one-shot classes (face identities);  shapes are  auxiliary labels (facial expression). With no constraints on embedding $f$ beyond a standard Triplet/ArcFace Loss, the auxiliary labels within each one-shot class may be distributed arbitrarily.  PDM  pulls examples within each one-shot class having the same auxiliary label close together. PDP tries to maintain, over all one-shot classes, a fixed distance $d_{a,b}$ between each \emph{pair} of examples with the same one-shot class whose auxiliary labels are $e_a$ and $e_b$, respectively. FBV is a stronger form of PDP: for each pair of auxiliary labels $e_b \neq e_a$, it fixes  $f(x_b) - f(x_a)$, where $e(x_b)=e_b$ and $e(x_a)=e_a$, to be a fixed vector. CE is a non-linear extension of FBV: using secondary function $g$, it estimates $f(x_b)$ as $g(f(x_a), e_a, e_b)$.}
    \label{fig:constraints}
\end{figure*}

\section{Examined Embedding Methods}
We propose new ways of improving the quality of embedding models by harnessing the geometric constraints imposed by emotion labels. We first describe the baseline approaches based on Triplet Loss and ArcFace loss functions.
Then we propose several novel loss functions that impose additional geometric structure in the embedding space: Pairwise Distance Minimization (PDM), Pairwise Distance Preservation (PDP) (similar to \cite{tsai2017improving}), 
Fixed Basis Vector (FBV), and Compositional Embeddings (CE) models.

\subsection{Triplet Loss}
A standard loss function for training an embedding model is the Triplet Loss (TL). Given 
three examples -- the anchor $x_a$, a positive example $x_p$ such that $c(x_a)=c(x_p)$, and
a negative example $x_n$ such that $c(x_n)\neq c(x_a)$ -- the loss is computed for each triplet as
\begin{eqnarray}
    \nonumber \lefteqn{L_\textrm{TL}(x_a,x_p,x_n) =} \\
    &&\|f(x_a) - f(x_p)\|_{2}^{2}-
    \|f(x_a) - f(x_n)\|_{2}^{2}+\alpha
\label{eqn:TL}
\end{eqnarray}
where $\alpha$ $\in$ $R$ is a margin hyperparameter.
The Triplet Loss encourages examples from the same class to be close together and encourages examples from different classes to be far apart. In Figure \ref{fig:constraints}, ``No constraints'' shows a hypothetical embedding space from such an approach:  Colors represent different one-shot classes (e.g.,  face identities), and shapes represent auxiliary labels. There is no explicit incentive for the embedding function to organize the auxiliary labels in any systematic way.
\subsection{ArcFace Loss}
The ArcFace loss as proposed in \cite{deng2019arcface} has achieved state-of-the-art results in various benchmark datasets. ArcFace inserts an angular margin geodesic distance between the sample and centers. ArcFace is based on the loss function
\begin{eqnarray*}
    \nonumber \lefteqn{L_\textrm{AF}(x_a, \overline{x}_p, \overline{x}_n)=} \\ 
    &&\textrm{GDis}(f(x_a), f(\overline{x}_p)) - 
    \textrm{GDis}(f(x_a), f(\overline{x}_n)) + \alpha
\label{eqn:AF}
\end{eqnarray*}
where $\textrm{GDis}$ is the geodesic distance (arc length) between two points embedded on a sphere, and $\overline{x}_p,\overline{x}_n$ are the centroids of the embeddings for persons $x_p$ (where $c(x_p)=c(x_a)$) and $x_n$ (where $c(x_n)\ne c(x_a)$), respectively.

\subsection{Pairwise Distance Minimization} \label{sec:PDM}
The first method we explored for enforcing more geometric structure on the embedding space using the auxiliary labels is the Pairwise Distance Minimization (PDM) Loss. It encourages all examples within each one-shot class that has the same auxiliary label to be close together. In this way, the PDM encourages the formation of ``mini-clusters'' within each one-shot cluster. One can thus view this loss function as a form of hierarchical clustering.  Figure \ref{fig:constraints} illustrates this idea. Given two examples: $x_a$ and $x_b$ such that $c(x_a)=c(x_b)$ and $e(x_a)=e(x_b)$, the PDM loss is computed for each pair: 
\begin{equation}
    L_\textrm{PDM}(x_a,x_b) = \|f(x_a) - f(x_b)\|_{2}^{2}
\label{eqn:PDM}
\end{equation}

We note that in contrast to multi-task learning (MTL) (see Section \ref{sec:MTL}) that defines \emph{implicit} relationships between embedding function and the emotion labels, PDM defines an \emph{explicit} relationship between the embedding representation and the auxiliary labels by imposing constraints on the loss function. Such explicit constraints may provide a better embedding space for one-shot learning compared to MTL.




\subsection{Pairwise Distance Preservation} \label{sec:PDP}
In the PDM loss we only explicitly encouraged the model to form clusters for \emph{individual} emotion labels. However, imposing constraints on how the multiple mini-clusters corresponding to different auxiliary labels should align with each other in the embedding space might help to better regularize the model. With the Pairwise Distance Preservation (PDP) loss we propose to encourage our model to maintain the same distance \emph{between} two auxiliary clusters across all one-shot classes. As shown in Figure \ref{fig:constraints}, we want auxiliary clusters corresponding to auxiliary labels $e_1$, $e_2$, and $e_3$ to have a similar distance across all one-shot classes. Given four examples: $x_1$, $x_2$, $x_3$ \& $x_4$ such that $c(x_1)=c(x_2)$, $c(x_3)=c(x_4)$, $c(x_1)\neq c(x_3)$ and $e(x_1)=e(x_3)$, $e(x_2)=e(x_4)$, $e(x_1)\neq e(x_2)$, the PDP loss is computed for each quadruple: 
\begin{eqnarray}
    \nonumber \lefteqn{L_\textrm{PDP}(x_1,x_2,x_3,x_4) =} \\ && \left(\|f(x_1) - f(x_2)\|_{2}^{2}-\|f(x_3) - f(x_4)\|_{2}^{2}\right)^2
\label{eqn:PDP}
\end{eqnarray}



\subsection{Fixed Basis Vector Separation} \label{sec:FBV}
With the PDP loss, we encourage the embedding space to generate mini-clusters based on the auxiliary labels with similar distances, but the relative positions of the mini-cluster centroids could still vary from one person to another. In contrast, the Fixed Basis Vector (FBV) loss that we propose forces the auxiliary clusters to align along a particular fixed vector across all the one-shot classes. 
In particular, we select one auxiliary class $e_1$ (say, a Neutral facial expression) as the ``origin'' each one-shot class. Then, for each other auxiliary label, we choose a unique Euclidean basis vector in the embedding space (e.g., $(1,0,\ldots,0)$ for Neutral to Anger, or $(0,1,0,\ldots,0)$ for Neutral to Sadness) as the desired vector between pairs of mini-clusters. In general,
given two examples $x_a$, $x_b$ such that $c(x_a)=c(x_b)$, $e(x_a)\neq e(x_b)$, and $v_{ab}$ is the unique fixed basis vector for the emotion pair $(e_a,e_b)$, the FBV loss is computed for the pair as 
\begin{equation}
    L_\textrm{FBV}(x_a,x_b) = \|f(x_b) - f(x_a)-v_{ab}\|_{2}^{2}
\label{eqn:FBV}
\end{equation}
One can also view the FBV loss as a combination of PDM and a stronger PDP since the FBV loss forces the individual auxiliary clusters to be close to each other.

Importantly, when using the FBV loss, we remove the constraint  -- which is commonplace with Triplet Loss and required for ArcFace loss -- that the embedding vectors must lie on a hypersphere. Instead, each embedding $y$ can be any point in Euclidean space.

\subsection{Compositional Embedding} \label{sec:CE}
The FBV loss encourages  $f$ to organize the embedded examples such that, for each one-shot class, the mini-cluster corresponding to $e_b$ can be reached from the mini-cluster corresponding to $e_a$ simply by adding a fixed basis vector $v_{ab}$. However, there may be other \emph{non-linear} mappings from one mini-cluster to another that more faithfully model the data and thereby yield an embedding space that separates the one-shot classes more accurately. With this goal in mind, we expand on the idea of \emph{compositional embeddings} (\cite{alfassy2019laso}; \cite{li2020compositional}) by introducing a compositional model $g$ to learn a non-linear map from one auxiliary mini-cluster to another in the embedding space. Through a joint training procedure, $g$ forces $f$ to align the auxiliary mini-clusters to be separable within each one-shot class while also potentially increasing the separability of the one-shot clusters themselves.

Suppose $x_a$ and $x_b$ are two examples from the same person that have different auxiliary labels, i.e., $c(x_a)=c(x_b)$, $e(x_a)\neq e(x_b)$.
We define our composition function $g$ and train it using the loss
\begin{eqnarray}
L_\textrm{CE}(x_a, x_b)=\|g(f(x_a), e(x_a), e(x_b)) - f(x_b)\|_2^2
\end{eqnarray}
This encourages $g$ to estimate $f(x_b)$ based on $f(x_a)$ and the auxiliary labels of these two examples. The CE method simplifies to FBV if we let $g(f(x_a), e(x_a), e(x_b))=f(x_a)+ v_{ab}$ for some fixed $v_{ab}$. In our implementation, $g$ consists of a 3 layer fully connected network (FCN(100)-ReLU-FCN(100)-ReLU-FCN(100)).

\subsection{Multitask Learning}
\label{sec:MTL}
One of the simplest methods to harness auxiliary information for improving latent representations is multi-task learning (MTL). Here, a single common latent layer is used to model multiple target variables (face identity and facial expression). The intuition for this idea is that training on multiple tasks helps to regularize the convolution layers, which could help perform better at the primary task of one-shot learning.

{\small
\bibliographystyle{ieee}
\bibliography{egbib}
}

\end{document}